\documentclass[conference]{IEEEtran}
\usepackage{times}
\usepackage{epsfig}
\usepackage{graphicx}
\usepackage{amsmath}
\usepackage{amssymb}
\usepackage[lined,algonl,ruled]{algorithm2e}
\usepackage{paralist}
\usepackage{color}
\usepackage{hyperref}
\usepackage{flushend}
\usepackage{geometry}
\usepackage{caption}
\usepackage{subcaption}
\usepackage{multirow}
\usepackage{balance}

\def\BibTeX{{\rm B\kern-.05em{\sc i\kern-.025em b}\kern-.08em
    T\kern-.1667em\lower.7ex\hbox{E}\kern-.125emX}}
\begin{document}

\title{Recognizing Magnification Levels\\in  Microscopic Snapshots\\
}

\author{\IEEEauthorblockN{Manit Zaveri$^1$, Shivam Kalra$^{1,2}$, Morteza
Babaie$^1$, Sultaan Shah$^2$, Savvas Damskinos$^2$,\\ Hany
Kashani$^1$, H.R. Tizhoosh$^{1,2}$}
\IEEEauthorblockA{$^1$\textit{Kimia Lab}, \textit{University of Waterloo},
Canada, https://kimialab.uwaterloo.ca/ \\ $^2$ Huron Digital Pathology,
St. Jacbos, ON, Canada\\ }}

\maketitle

\begin{abstract}
Recent advances in digital imaging has transformed computer vision and machine
learning to new tools for analyzing pathology images. This trend could automate some
of the tasks in the diagnostic pathology and elevate the pathologist workload.
The final step of any cancer diagnosis procedure is performed by the expert
pathologist. These experts use microscopes with high level of optical
magnification to observe minute characteristics of the tissue acquired
through biopsy and fixed on glass slides. Switching between different
magnifications, and finding the magnification level at which they identify the presence
or absence of malignant tissues is important. As the majority of pathologists
still use light microscopy, compared to digital scanners, in many instance a
mounted camera on the microscope is used to capture snapshots from significant
field-of-views. Repositories of such snapshots usually do not contain the
magnification information. In this paper, we extract deep features of the images
available on TCGA dataset with known magnification to train a classifier for
magnification recognition. We compared the results with LBP, a well-known
handcrafted feature extraction method. The proposed approach achieved a mean
accuracy of 96\% when a multi-layer perceptron was trained as a classifier.
\end{abstract}


\section{Introduction}
\label{sec:intro}
A biopsy followed by specimen preparation in laboratory and a subsequent
microscopic examination by a trained pathologist is necessary for a definitive
diagnosis of any type of cancer and many other diseases. Hematoxylin and eosin (H\&E)
staining is applied to thin cuts of the biopsy sample to visualize the structural
patterns and any distortion of the tissue. To differentiate benign from
malignant cells and to extract the distinctive cell features, the pathologist generally observes the tissue at several magnification levels to gain a more comprehensive
understanding of the specimen. The size of the area under observation decreases
with increase in magnification, allowing experts to view the enlarged tissue and
observe the minute characteristics relevant for diagnosis. Some microscopes are
equipped with a mounted camera that is used to capture snapshots from the glass
slide. Pathologists create snapshots of the tissues and save them for future
reference in reports or for research purposes. These snapshots usually miss some crucial
information such as the magnification level which is required for many classification tasks. Therefore, the use of these snapshots for future
research is quite limited particularly if there are large and diverse repositories
of such snapshots of different organs and diseases. This limitation is the main
motivation for this work~\cite{clark1994microscope,winterot2005microscope}.

The trade-off between the superior performance and time complexity in computer vision research is mainly noticeable when it comes to feature extraction. There are several algorithms for feature extraction like SIFT, HOG and LBP which are based on handcrafted design. In contrast, there are also deep learning methods using convolutional neural networks (CNN) like VGG, ResNet and Incecption which are based on learning from data. The modern AI algorithms have been intensively investigated recently~\cite{tizhoosh2018artificial}. Deep networks can learn distinctive image features, while traditional algorithms implement a series of functions on the image to extract  important characteristics. These features are being deployed for various classification and content-based image retrieval tasks~\cite{kalra2019yottixel}. Both techniques, handcrafted and learning-based, have proven to perform well for different applications although deep features are reported to generally be much more expressive ~\cite{8285162,otalora2018image}. Recent publications have investigated this problem to be generic in histopathology domain. Several approaches for classification of malignancy in breast cancer images, for instance, have been performed using various techniques. Bayramoglu et al.~\cite{bayramoglu2016deep} utilized deep features while Gupta et al.~\cite{gupta2017breast} used color texture features and evaluated the influence of magnification on classification model to identify malignancy. Otalora et al.~\cite{otalora2018image} focused only on magnification level and implemented a CNN-based regression to find the magnification level using multiple open access datasets.

This paper is organized as follows: The feature extractors and classification
algorithms are briefly discussed in~\autoref{sec:Background}. The dataset is introduced
in ~\autoref{sec:Dataset}.~\autoref{sec:Methodology} explains the methodology
used in the study. Experiments and results are discussed in~\autoref{sec:exp and
results}. Finally~\autoref{sec:con} concludes the paper.

\section{Background}
\label{sec:Background}

\subsection{Feature Extraction}
\subsubsection{Local Binary Pattern (LBP)} LBP is a conventional feature
extraction algorithm. Ojala and Pietikainen’s research in multi-resolution
approach~\cite{ojala2002multiresolution} proved LBP's useful application in text
identification~\cite{jung2010local} and histopathology
classification~\cite{caicedo2009histopathology,erfankhah2019heterogeneity}. A
combined approach of LBP descriptor with histogram of Oriented Gradients (HOG)
descriptor was used by Wang ~\cite{wang2009hog} to improve the detection performance.

There are many LBP variations available which can address the effect of rotational 
invariance uniformity on neighborhood pixels~\cite{nigam2019local}. A brief
description of most recent applications of LBP are discussed in a
survey~\cite{kas2019survey}.

\subsubsection{Pre-Trained Deep Nets} Pre-trained deep networks can be
employed to transfer knowledge from one domain to another domain. For instance,
DenseNet121 is a deep convolutional neural network proposed by Huang et
al.~\cite{huang2017densely} that has been applied to many different problems.
Recent publications emphasize the capabilities of artificial neural networks and
their performance in several classification and search-based
tasks~\cite{srinidhi2019deep,gordo2016deep}.

\subsection{Classifiers}
There is a large body of literature on different algorithms for data
classification. We restrict ourselves to the following two approaches mainly
because of our previous experience with these methods.

\subsubsection{Artificial Neural Networks (ANNs)} ANNs are classifiers with the
capability to simulate the function of the human brain at a very small scale and
for a given task. They are commonly used as robust classifiers in machine
learning when a large body of labelled data is available for training. The
network mainly consists of three primary layers: the first layer represents
input neurons; the last layer represents output neurons; and hidden layers
consist of a series of weighted layers which minimize the error between actual
output and predicted output. It is difficult to interpret the trajectory of ANNs
toward output to understand the rational behind the decision.

When we talk about ANNs we generally mean shallow networks (less than 5 layers),
in contrast to convolutional neural networks like DenseNet, ResNet and VGG-19
that are deep networks with many hidden layers.

\subsubsection{Random Forests (RF)} Random Forest (RF) classifiers are multi-way
decision trees with some randomization used to grow each tree as a potential
solution. The leaf nodes represent the posterior distribution of each image
class. Internal nodes contain a test split based on the maximum information
gained from the feature subspace. Bosch et al. implemented image classification
using random forests beating state-of-the-art results~\cite{bosch2007image}.

\section{Dataset}
\label{sec:Dataset}
\begin{figure*}[htb]
  \centering
  \includegraphics[width=\textwidth]{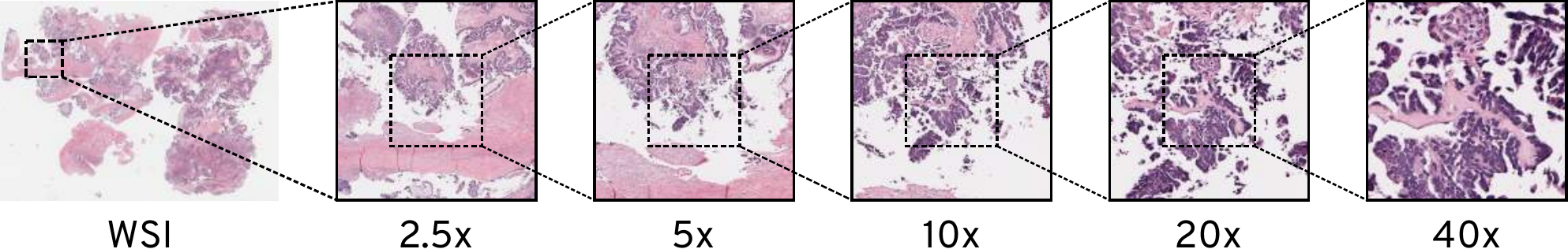}
  \caption{Illustration of patches from different magnification levels extracted
    from a WSI. All the patches are centered around the same coordinates (images
    re-scaled for convenient visualization).}
    \label{fig:maglevels}
\end{figure*}

We used publicly available data from The Cancer Genome Atlas (TCGA). An
important characteristic of this dataset is the available objective power of the
whole slide images, which represents different magnification levels. Other publicly available
datasets like PMC do not contain magnification information, making them
unsuitable for this study. Each image was stored at various magnification levels
and in a pyramid structure. The subset from the original dataset was created
using an indexing algorithm, which would randomly index one whole slide image
(WSI) at a time. If the objective power(i.e., the magnification of base layer)
of 40x or 20x was absent, we discarded that WSI. In total, we gathered 29,596
WSIs for creating our magnification specific dataset. For these WSI files we
randomly selected the coordinates of 5 points, read the image at respective
magnifications from that specific coordinates and took a snapshot at each point.
This yielded us the total of 693,518 patches, consisting of 147,477 patches at
2.5x,5x,10x and 20x and 103,611 at 40x. A region of a sample snapshot at different 
magnification levels is represented in \autoref{fig:maglevels}.

\section{Methodology}
\label{sec:Methodology}
For classifying the magnification, two models are trained based on different
features, one CNN-based approach using DenseNet121 and one conventional
algorithm, namely Local Binary Patterns (LBP). The vectors obtained by each feature extractor is
considered as an input for classification models. These features are
independently fed into ANNs and RF to train them for magnification recognition.
The performance of classification is evaluated using a 5-fold cross-validation
with 80\%-20\% split for training and testing, respectively. The accuracy
provides a performance index for correct classification, kappa score provides
empirical probability of agreement associated to each label and the F1-score
provides the arithmetic mean of precision and
recall~\cite{otalora2018image,scikit-learn}.

\subsection{Feature Extraction with LBP} The feature vectors obtained through
the application of LBP are conserved in histograms. We used rotationally
invariant LBP with three parameter settings for radius $r$ and neighbors $n$.
For radius $r=1$ and neighbours $n=8$ we got features with 10 dimensions (bins).
Radius $r=2$ and neighbours $n=16$ resulted in features with $16$ bins, and
features with $26$ bins were the result of radius $r=3$ and neighbours $n=24$.

\subsection{Deep Features using DenseNet121}
Using pre-trained features for histopathology images has been the focus of
attention in recent literature \cite{kieffer2017convolutional}. Image features
were extracted from the last average pooling layer in DenseNet121. Before
passing through the feature extractor, we pre-processed all images by making the
mean zero. The input tensor flows through convolutional feature layer to capture
low-level information consisting of shapes from histopathology images. This was
followed by a series of 4 dense blocks. These dense blocks have 6, 12, 16 and 24
layers of batch normalization, reLu and 2D- convolutional layers stacked
alternatively. The feature vector from the last average pooling layer contained
1,024 dimensions which were passed to the classifier to represent the
corresponding image magnification.

In total, we had 4 individual feature sets for our dataset. One may argue that
using ANNs is considered a fine-tuning technique, but for all our experiments we
treated classification methods separate from feature extraction models.

\section{Experiments And Results}
\label{sec:exp and results}

\subsection{ANNs}
The shallow classifier that used deep features excelled in performance. The
accuracy was observed at 96.1\% which was achieved using 2 hidden layers and one
fully connected layer between input and output layers. The hidden layers had 512
neurons with batch normalization and a dropout rate of 0.5 followed by other
fully connected layer of 256 neurons and ReLU activation. This finally was
connected to output layer of 5 neurons which was activated using softmax. For
every ANN we matched the input layer neurons with the dimensions of the input
feature vector. For LBP features, ANNs achieved an accuracy of 68.4\%, kappa
score of 0.603 and f1-score of 0.676 for one split. From \autoref{tab:results}
we can easily identify the higher performance of deep features compared to
handcrafted LBP features. However, there is a trade-off between real-time high
performance and time complexity plus the computational power utilized for
extracting deep features. The confusion matrix in~\autoref{fig:confusion_matrix}
shows that images at a magnification level of 5x are mostly confused with class
of images at 2.5x and 10x magnification. The individual class accuracy was 91\%
for 5x images. While images at 2.5x, 20x and 40x were separated excellently with
misclassification rate of less than 1\%.

\subsection{Random Forests}
\begin{figure}[htb]
\centering
\begin{subfigure}[b]{0.38\textwidth}
\centering
		\includegraphics[width=\textwidth]{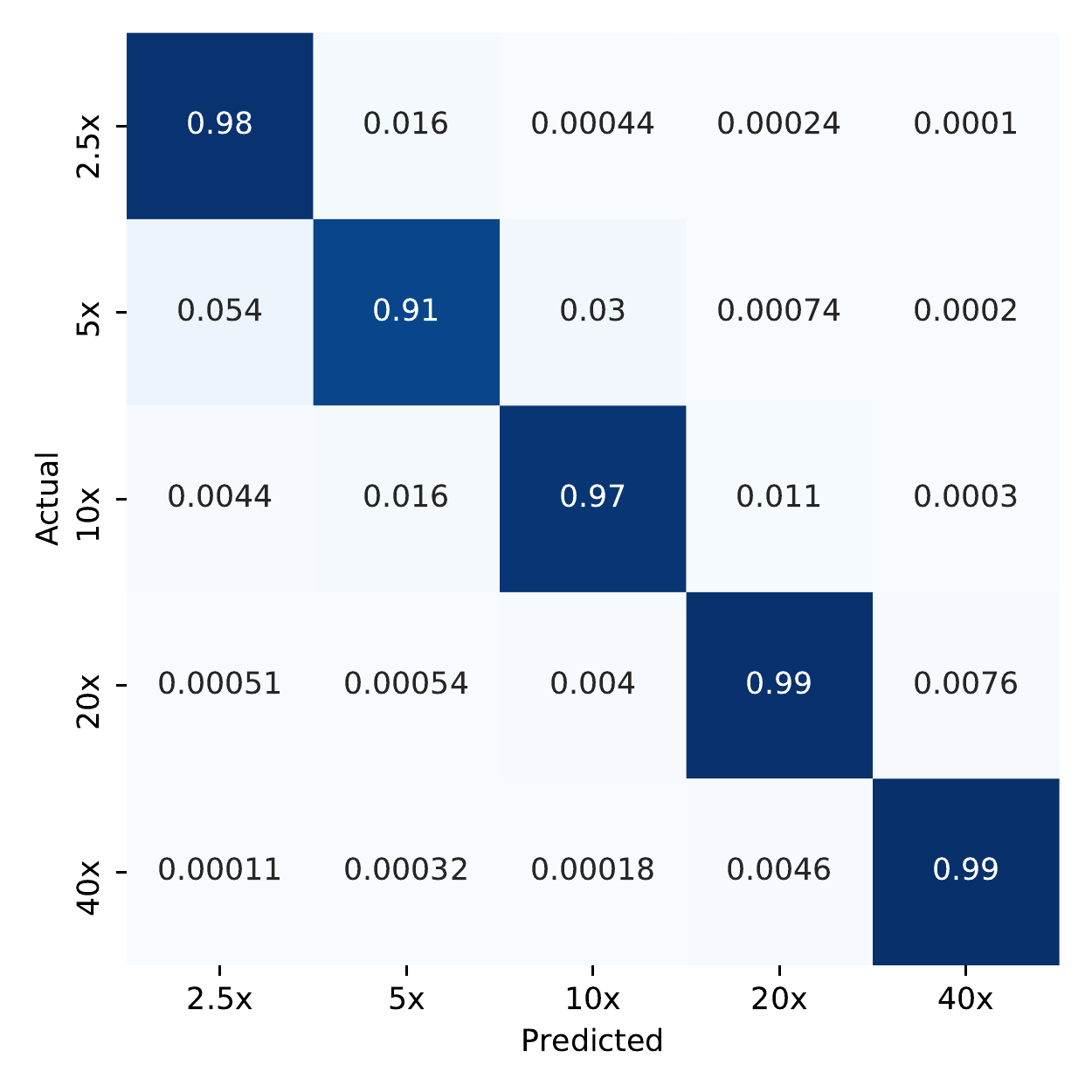}
		\caption{ANN Classifier}
		\label{fig:mlp_cf}
	\end{subfigure}\\
	\vspace*{-1mm}
	\begin{subfigure}[b]{0.38\textwidth}
	\centering
		\includegraphics[width=\textwidth]{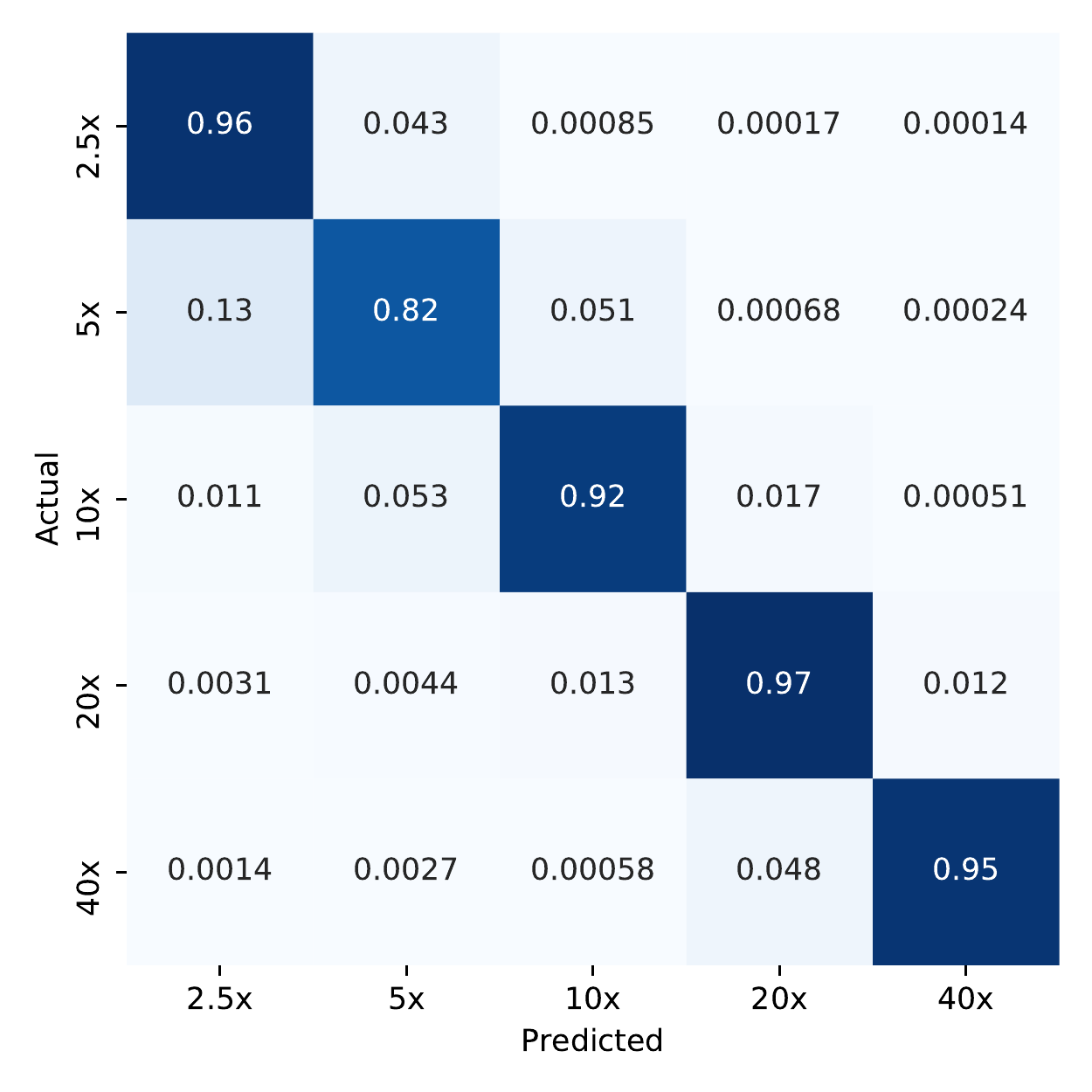}
		\caption{RF Classifier}
    	\label{fig:rf_cf}
	\end{subfigure}
	\caption{Confusion matrix of the two classifiers.}
\label{fig:confusion_matrix}
\end{figure}

We searched for best parameters using random searchCV and grid searchCV from
Scikit Learn library~\cite{scikit-learn}. We used 1000 estimators for training
the classifier, with a maximum depth of 50 splits for individual trees. The
final label assignment was determined through majority voting from each tree.
The reported accuracy for classification using deep features was 96\% with kappa
score of 0.95 and F1-score of 0.96 while using LBP, the best accuracy obtained
by RF classification model was 68.4\% with kappa score of 0.60 and F1-score of
0.67. All results using RF classifier are listed in \autoref{tab:results}.
Comparing the confusion matrices in ~\autoref{fig:confusion_matrix} a and b, we
can see that classifiers using deep features are more prone to confusing 5x
images compared to shallow classifiers. While classes 20x and 40x are also more
prone to misclassification when RFs are implemented.

\begin{table*}[t]
  \begin{center}
    \scalebox{0.9}{
\begin{tabular}{|c|c|c|c|c|c|c|c|l}
\cline{1-8}
\multirow{2}{*}{Features}& \multirow{2}{*}{Folds}& \multicolumn{3}{ |c| }{Random Forests}&\multicolumn{3}{|c|}{Shallow Classifier} \\ \cline{3-8}
 & & Acc & Kappa & F$1$-Score & Acc & Kappa & F$1$-Score \\ \cline{1-8}
{\multirow{6}{*}{DF} } 
&{Fold1} & 0.881 & 0.849 & 0.881 & 0.935 & 0.917 & 0.935 \\
&{Fold2} & 0.879 & 0.849 & 0.879 & 0.96 & 0.95 & 0.96 \\
&{Fold3} & 0.916 & 0.849 & 0.915 & 0.971 & 0.963 & 0.971 \\
&{Fold4} & 0.924 & 0.905 & 0.924 & 0.972 & 0.966 & 0.973 \\
&{Fold5} & 0.921 & 0.901 & 0.921 & 0.969 & 0.962 & 0.969 \\
&{All Folds} & 0.904 $\pm$ 0.01 & 0.879 $\pm$ 0.025 & 0.904 $\pm$ 0.01& $\mathbf{0.961 \pm 0.01}$ & $\mathbf{0.951 \pm 0.01}$ &$\mathbf{0.961 \pm 0.01}$ \\ 
\cline{1-8}

{\multirow{6}{*}{LBP$_1$} } 
&{Fold1} & 0.520 & 0.392 & 0.509 & 0.514 & 0.382 & 0.501 \\
&{Fold2} & 0.612 & 0.513 & 0.604 & 0.601 & 0.499 & 0.588 \\
&{Fold3} & 0.649 & 0.560 & 0.643 & 0.598 & 0.496 & 0.583 \\
&{Fold4} & 0.684 & 0.605 & 0.679 & 0.646 & 0.557 & 0.631 \\
&{Fold5} & 0.684 & 0.605 & 0.678 & 0.655 & 0.568 & 0.646 \\
&{All Folds} & 0.629 $\pm$ 0.06 & 0.535 $\pm$ 0.07 & 0.622 $\pm$ 0.06 & 0.602 $\pm$ 0.05 & 0.50 $\pm$ 0.06 & 0.589 $\pm$ 0.05 \\ \cline{1-8}

{\multirow{6}{*}{LBP$_2$} } 
&{Fold1} & 0.516 & 0.389 & 0.508 & 0.471 & 0.339 & 0.458 \\
&{Fold2} & 0.582 & 0.476 & 0.571 & 0.565 & 0.453 & 0.556 \\
&{Fold3} & 0.626 & 0.532 & 0.618 & 0.587 & 0.480 & 0.578 \\
&{Fold4} & 0.656 & 0.569 & 0.649 & 0.613 & 0.516 & 0.607 \\
&{Fold5} & 0.654 & 0.567 & 0.647 & 0.628 & 0.534 & 0.618 \\
&{All Folds} & 0.606 $\pm$ 0.05 & 0.506 $\pm$ 0.06 & 0.598 $\pm$ 0.05 & 0.572 $\pm$ 0.05 & 0.464 $\pm$ 0.06 & 0.563 $\pm$ 0.05 \\ \cline{1-8}

{\multirow{6}{*}{LBP$_3$} } 
&{Fold1} & 0.546 & 0.425 & 0.536 & 0.507 & 0.375 & 0.493 \\
&{Fold2} & 0.609 & 0.510 & 0.599 & 0.582 & 0.476 & 0.574 \\
&{Fold3} & 0.649 & 0.561 & 0.643 & 0.634 & 0.542 & 0.622 \\
&{Fold4} & 0.682 & 0.603 & 0.676 & 0.668 & 0.584 & 0.656 \\
&{Fold5} & 0.676 & 0.595 & 0.671 & 0.640 & 0.551 & 0.631 \\
&{All Folds} & 0.632 $\pm$ 0.05 & 0.538 $\pm$ 0.06 & 0.624 $\pm$ 0.05 & 0.606 $\pm$ 0.05 & 0.505 $\pm$ 0.07 & 0.595 $\pm$ 0.05 \\ \cline{1-8}
\end{tabular}
}
\caption{Results for deep features (DF) and LBP with 3 parameter settings.}
\label{tab:results}
\end{center}
\end{table*}

\section{Conclusions}
\label{sec:con}

In this paper, two feature extraction methods were used to train two classifiers
for discrete classification of magnification levels in microscopic images. We
used the TCGA dataset to construct a large dataset for training and testing. The
feature extraction was done from raw images without any pre-processing or
segmentation. A shallow classifier with deep features obtained from DenseNet121
had the best performance in terms of accuracy (96\%), kappa (0.95) and F1-score
(0.96). The score achieved by our model outperformed existing state-of-the-art
regression method with 11\% improved kappa score. Because the TCGA dataset
consists of various categories of images and several organs, we assume the model
offers good generalization for magnification recognition providing a basic pre-processing step for all digital pathology image processing to attain better performance. 

Microscopic snapshots may have a different image quality than patches from whole slide images. Investigating this difference will be subject of future work. 

\bibliography{mag_learning}

\end{document}